**Title:** Leveraging Fine-Tuned Large Language Models for Interpretable Pancreatic Cystic Lesion Feature Extraction and Risk Categorization

**Authors:** Ebrahim Rasromani, MS · Stella K. Kang, MD, MS · Yanqi Xu, MS · Beisong Liu, BS · Garvit Luhadia, MS · Wan Fung Chui, MD · Felicia L. Pasadyn, MA · Yu Chih Hung, MD · Julie Y. An, MD · Edwin Mathieu, MD · Zehui Gu, MS · Carlos Fernandez-Granda, PhD · Ammar A. Javed, MD, PhD· Greg D. Sacks, MD, PhD, MPH · Tamas Gonda, MD · Chenchan Huang, MD* · Yiqiu Shen, PhD*

**Author Affiliations:**

- From the Center for Data Science (E.R., Y.X., B.L., G.L., Y.S.), and the Department of Mathematics, Courant Institute of Mathematical Sciences (C.F.-G.), New York University, New York, NY, USA; Department of Radiology (S.K., Z.G.), Columbia University Irving Medical Center (S.K.K., Z.G.), New York, NY, USA; Department of Radiology (W.F.C., F.L.P., J.Y.A., E.M., Y.C.H., C.H., Y.S.), Department of Surgery (A.A.J., G.D.S.), and Department of Medicine (T.G.), NYU Grossman School of Medicine, New York, NY, USA; Department of Radiology (J.Y.A.), University of California San Diego, La Jolla, CA, USA;
- **Address correspondence to** Y.S. (email: Yiqiu.Shen@nyulangone.org).
- * Equal contribution senior authors

**Abbreviation:** LLM = Large Language Model, CoT = Chain of Thought, QLoRA = Quantized Low-Rank Adaptation, PCL = Pancreatic Cystic Lesion, WF = Worrisome Features, HRS = High-Risk Stigmata, JSON = JavaScript Object Notation

**Summary Statement**: Fine-tuned open-source LLMs with chain-of-thought supervision were developed to extract pancreatic cystic lesion features from CT/MRI reports. Risk categories were assigned based on the extracted features, achieving performance comparable to GPT-4o and agreement on par with radiologists.

**Key Results**:

- Fine-tuned open-source large language models accurately extracted pancreatic cyst features from full CT and MRI radiology reports and assigned risk categories, achieving 97–98% average exact match accuracy across all features.
- Performance of fine-tuned open-source models matched that of GPT-4o in both feature extraction and risk categorization, while offering substantially lower inference costs at scale.
- In a reader study, model-assigned risk categories demonstrated near-perfect agreement with expert radiologists (Fleiss' kappa: 0.893–0.897), supporting the clinical reliability of structured outputs generated through chain-of-thought reasoning.

**Abstract**

**Background:** Manual extraction of pancreatic cystic lesion (PCL) features from radiology reports is labor-intensive, limiting large-scale studies needed to advance PCL research.

**Purpose:** To develop and evaluate large language models (LLMs) that automatically extract PCL features from MRI/CT reports and assign risk categories based on guidelines.

**Materials and Methods**: We curated a training dataset of 6,000 abdominal MRI/CT reports (2005–2024) from 5,134 patients that described PCLs. Labels were generated by GPT-4o using chain-of-thought (CoT) prompting to extract PCL and main pancreatic duct features. Two open-source LLMs were fine-tuned using QLoRA on GPT-4o-generated CoT data. Features were mapped to risk categories per institutional guideline based on the 2017 ACR White Paper. Evaluation was performed on 285 held-out human-annotated reports. Model outputs for 100 cases were independently reviewed by three radiologists. Feature extraction was evaluated using exact match accuracy, risk categorization with macro-averaged F1 score, and radiologist–model agreement with Fleiss' kappa.

**Results**: CoT fine-tuning improved feature extraction accuracy for LLaMA (80% to 97%) and DeepSeek (79% to 98%), matching GPT-4o (97%). Risk categorization F1 scores also improved (LLaMA: 0.95; DeepSeek: 0.94), closely matching GPT-4o (0.97), with no statistically significant differences. Radiologist inter-reader agreement was high (Fleiss' κ = 0.888) and showed no statistically significant difference with the addition of DeepSeek-FT-CoT (κ = 0.893) or GPT-CoT (κ = 0.897), indicating that both models achieved agreement levels on par with radiologists.

**Conclusion**: Fine-tuned open-source LLMs with CoT supervision enable accurate, interpretable, and efficient phenotyping for large-scale PCL research, achieving performance comparable to GPT-4o.

## Introduction

Pancreatic cystic lesions (PCLs) are highly prevalent, approaching 50% in 60-year-old individuals in a population-based abdominal MRI study, and increasing progressively with each subsequent decade (1). A small percentage progress to malignancy, defined as high-grade dysplasia and invasive carcinoma (2). Multiple specialty society guidelines recommend long-term imaging surveillance to monitor for evidence of malignant transformation (3–5). However, studies have shown that none of these guidelines achieve both high sensitivity and specificity for detecting malignancy (6,7). There is therefore a critical need to better understand the natural history of PCLs and improve risk categorization. Progress is currently hindered by the low prevalence of high-risk PCL cases—those with worrisome features (WF) and high-risk stigmata (HRS) (4).

Addressing key research questions in PCLs requires large-scale and ideally multi-institutional radiology datasets. To enable such studies, there is an unmet need for efficient, scalable methods to extract structured PCL-related information from radiology reports. Manual review is time-consuming and error-prone (8). Recent advances in large language models (LLMs) have demonstrated promise in natural language processing (NLP) tasks across healthcare applications (9). While prior studies have shown the feasibility of LLM for automated PCL feature extraction, they are limited by narrow feature coverage, focus only on the largest cyst instead of the most concerning one, and use only segmented sections of the report as input rather than the full report (10–12). In addition, existing approaches frequently rely on closed-source LLMs, which are costly, raise privacy concerns, and offer limited control over decoding and reproducibility. Conversely, open-source LLMs without domain-specific fine-tuning often underperform due to limited adaptation to medical language and clinical reasoning. Moreover, these methods may face scalability challenges in high-volume settings, where processing millions of reports efficiently requires careful tradeoffs between inference cost, latency, and infrastructure control (13).

To address these limitations, we developed fine-tuned, open-source LLMs that extract a comprehensive set of PCL features from radiology reports and assign risk categories based on clinical guidelines. Unlike prior methods, our models reason over the entire report to identify the most clinically significant PCL and generate interpretable outputs through chain-of-thought (CoT) reasoning (14). This approach reduces reliance on proprietary systems for inference, eliminates the need for manual label annotation for training, and supports cost-efficient, scalable deployment. In this study, we systematically evaluated the performance of closed-source, open-source, and fine-tuned open-source LLMs for structured PCL feature extraction and risk categorization, with an emphasis on interpretability and high-throughput inference. Figure 1 provides an overview of the study.

## Materials and Methods

This retrospective HIPAA-compliant study was approved by an institutional review board (IRB i24-0111) with a waiver of informed consent. The study design and reporting comply with the CLAIM guideline.

*Data Collection and Preparation*

Patients who underwent abdominal MRI or CT between 2005 and 2024 at NYU Langone Health with PCLs were eligible for inclusion. A keyword search of radiology reports for "pancreatic cystic lesions," "pancreatic cysts," "side-branch IPMN, "intraductal papillary mucinous neoplasm", and "branch-duct IPMN" identified 30,122 patients (55,552 reports). From this cohort, 6,469 reports from 5,615 patients (median age: 70; range: 18–103; 59.7% female) were randomly selected. Reports included both structured (template-based) (15) and unstructured (free-text) formats . Patients were randomly split into training (n=5,153), validation (n=181), and test (n=281) sets, ensuring all exams from a given patient were in a single subset. Model selection was based on the validation set; the test set was used exclusively for final evaluation. Dataset characteristics are summarized in Table 1.

Training labels were generated using GPT-4o with CoT prompting to extract key PCL features, including cyst characteristics (size, growth, morphology, ductal communication, wall and septation characteristics, mural nodules) and main pancreatic duct features (caliber, dilation, stricture), as well as differential diagnoses and presence of pancreatitis (Table S1, Appendix I). For evaluation, 285 test set cases were manually annotated by two trainees (one radiology resident, W.F.C.; one medical student; F.L.P.) under supervision of an abdominal radiologist (C.H.), focusing on the dominant cyst and main pancreatic duct features.

*PCL Feature Extraction*

We framed PCL feature extraction as a structured text completion task. Each model received the full radiology report and generated a structured JSON of extracted features in response to a prompt specifying extraction instructions, provided in full in Appendix II. We evaluated seven LLM-based approaches that varied by base model (GPT-4o, LLaMA, or DeepSeek), prompting strategy (standard or CoT), and decoding method (greedy, beam search, or temperature sampling). Exact match accuracy was used as the primary metric for feature extraction.

Closed-source GPT-4o models were evaluated with and without CoT prompting. In the CoT setting, the model generated intermediate reasoning traces that quoted relevant report text and verbalized the reasoning that led to each final feature value, as illustrated in Appendix III. Open-source LLaMA-3.1-8B-Instruct and DeepSeek-R1-Distill-Llama-8B were fine-tuned using Quantized Low-Rank Adaptation (QLoRA) on GPT-4o-generated CoT data (16–18).

We adopt the following naming convention for models: GPT and GPT-CoT denote GPT-4o with standard and CoT prompting, respectively; LLAMA and DeepSeek refer to zero-shot prompted open-source models; LLAMA-FT indicates a fine-tuned model trained to output structured JSON only; and LLAMA-FT-CoT and DeepSeek-FT-CoT refer to models fine-tuned with CoT supervision. An eighth variant, DeepSeek-FT-CoT-SC, uses self-consistency decoding for scalable deployment and is presented in Appendix V (19). Implementation details, prompt examples, training configuration, and decoding strategies are provided in Appendices II–IV.

*PCL Risk Categorization*

LLM-extracted features were mapped to PCL risk categories using an institutional guideline which is based on the 2017 American College of Radiology (ACR) White Paper (4) and the 2017 International Association of Pancreatology (IAP)/Fukuoka guidelines (20). This guideline was operationalized using a framework published by the PRECEDE consortium (15). Criteria for each risk category are summarized in Table 2. In the test set, 52% of cases involved low-risk PCLs, 43% involved PCLs with WF or HRS, and the remaining 5% did not describe PCLs. Model performance was evaluated on the test set using the macro-averaged F1 score.

*Reader Study*

We conducted a reader study to assess agreement between LLM-assigned PCL risk categories and radiologist interpretations. Three abdominal radiologists (two fellows, J.A. and E.M.; one attending, Y.H.) independently reviewed 100 test set cases. For each case, readers were shown the radiology report, model-extracted features, and model-assigned risk category. Readers indicated whether they agreed with the model's risk category and provided their own final categorization based solely on the report text. Each radiologist reviewed outputs from both GPT-CoT and DeepSeek-FT-CoT, blinded to model identity. Agreement was quantified using percent agreement and Cohen's kappa per reader. Fleiss' kappa was first computed among the three radiologists, then recomputed with the LLM as a fourth reader to assess whether overall agreement was maintained (21). Of the 100 reports, 87 were unstructured and 13 were structured (15).

*Statistical Analysis*

Statistical comparisons of LLM accuracy in extracting PCL features were conducted using Wilcoxon signed-rank tests applied to paired feature-level exact match accuracies. One-sided tests evaluated fine-tuned vs. non-fine-tuned open-source models and GPT-4o with vs. without CoT prompting. A two-sided test compared GPT-CoT to fine-tuned open-source models under the null hypothesis of equivalent performance.

Permutation tests were used to compare model performance on individual features (one-sided) and macro-averaged F1 scores for PCL risk categorization (two-sided). For the reader study, two-sided permutation tests evaluated whether Fleiss' kappa significantly changed when the LLM was added as a fourth reader, simulating the null hypothesis that the LLM is exchangeable with a radiologist. Fleiss' kappa values were interpreted using established thresholds: >0.81 = almost perfect, <0 = poor agreement (22). Confidence intervals were obtained via bootstrapping.

Correction was applied only to hypothesis families with more than two comparisons. Holm–Bonferroni correction was used to control the family-wise error rate at an overall significance level of 0.05. All other p-values are unadjusted. Full implementation details (e.g., statistical libraries, resampling settings) are provided in Appendix VI.

**Results**

We evaluated model performance across five dimensions relevant to clinical deployment and interpretability: PCL feature extraction accuracy, PCL risk categorization performance, error and hallucination analyses, radiologist agreement assessment, and cost-efficiency. Results for seven approaches appear in the main text; the eighth variant (DeepSeek-FT-CoT-SC) is detailed in Appendix V.

*PCL Feature Extraction Accuracy*

We first evaluated model performance on structured PCL feature extraction. As shown in Figure 2, fine-tuned models outperformed their non-fine-tuned counterparts in exact match accuracy across all features. LLAMA-FT-CoT improved average accuracy from 80% (95% CI: 79–81%) to 97% (95% CI: 97–98%; p<0.001), and DeepSeek-FT-CoT from 79% (95% CI: 77–81%) to 98% (95% CI: 97–98%, p<0.001). Fine-tuning with CoT supervision was particularly beneficial for features requiring multi-step reasoning. For example, on the Time Interval feature, LLAMA-FT-CoT achieved 91% accuracy (95% CI: 87–94%) compared to 85% (95% CI: 81–89%) without CoT (p=0.0063).

GPT-CoT also demonstrated an improvement (p=0.0013) over GPT, with average accuracy increasing from 95% (95% CI: 94–95%) to 97% (95% CI: 97–98%), confirming the utility of CoT even in closed-source models. Despite this improvement in GPT-CoT performance, there was no significant difference in paired feature-level accuracies between GPT-CoT and the fine-tuned open-source models (p = 1.00 vs. LLAMA-FT, p = 1.00 vs. LLAMA-FT-CoT, and p = 1.00 vs. DeepSeek-FT-CoT). These results highlight a central finding: with targeted fine-tuning, open-source models matched GPT-CoT in feature extraction performance.

*PCL Risk Categorization Performance*

Figure 3 presents model performance in PCL risk categorization, highlighting three key patterns. First, F1-scores were comparable across models—GPT-CoT achieved 0.97 (95% CI: 0.93–0.99), LLAMA-FT 0.95 (95% CI: 0.91–0.98), DeepSeek-FT-CoT 0.94 (95% CI: 0.90–0.98), and LLAMA-FT-CoT 0.93 (95% CI: 0.89–0.97)—with no statistically significant differences between GPT-CoT and LLAMA-FT-CoT (p = 0.1548), LLAMA-FT (p = 0.4322), or DeepSeek-FT-CoT (p = 0.4322). These findings highlight a key result: fine-tuned open-source models can achieve risk categorization performance on par with GPT-CoT, a state-of-the-art proprietary model. Second, recall consistently exceeded precision across all models, indicating a tendency to prioritize recall over precision. For example, GPT-CoT achieved a precision of 0.96 (95% CI: 0.92–0.99) and a recall of 0.97 (95% CI: 0.94–0.99); DeepSeek-FT-CoT achieved 0.93 (95% CI: 0.88–0.97) precision and 0.96 (95% CI: 0.93–0.99) recall; and LLAMA-FT-CoT had 0.91 (95% CI: 0.86–0.96) precision and 0.95 (95% CI: 0.92–0.98) recall. Third, while recall remained stable across higher-risk categories, precision declined most notably in the highest-risk group. For example, LLAMA-FT maintained stable recall across all risk categories (0.96–0.98), but its precision declined from 0.99 in low-risk cases (category 1) to 0.83 in high-risk cases (category 3), illustrating a 16-point drop for the most clinically significant group. This consistent emphasis on recall may reflect cautious tendencies—possibly shaped by reinforcement learning with human feedback (RLHF)—that favor identifying high-risk cases even at the expense of more false positives (23).

*Error and Hallucination Analyses*

Error and hallucination analyses are critical for understanding not just how models fail, but why—revealing reasoning flaws, misinterpretations, and whether outputs are faithfully grounded in the source text. Figure 4 presents results from the error and hallucination analyses for all CoT models. We reviewed each model's reasoning trace to understand how it derived values for specific features and categorized errors using the taxonomy in Table S2 (Appendix I). The most common error types were object identification errors (e.g., misidentifying the most worrisome PCL or miscounting the number of PCLs) and clinical reasoning errors (e.g., misapplying clinical logic when interpreting radiologic descriptors). For example, as an object identification error, DeepSeek-FT-CoT marked Number of Cysts Measured as one, despite the report describing two 2 mm cysts—one in the proximal and one in the distal portion of the pancreatic tail. As a clinical reasoning error, GPT-CoT incorrectly marked a cyst as having thickened septations based on the impression phrase "progressive septal enhancement," even though the findings explicitly described "thin converging septations with progressively delayed enhancements". GPT-CoT made fewer calculation errors (3.9%) compared to DeepSeek-FT-CoT (6.3%) and LLAMA-FT-CoT (16.0%). For instance, LLAMA-FT-CoT miscalculated the time interval between two study dates—10/27/2017 and 5/7/2018—reporting 222 days (7 months) instead of the correct 192 days (6 months), resulting in an incorrect Time Interval value.

In parallel, we conducted a hallucination analysis to assess whether models faithfully quoted text from the original reports as evidence. The reasoning trace consisted of two components: an observation, which quoted relevant text from the report pertaining to the feature being extracted, and a reasoning step, which built on the observation by verbalizing the thought process leading to the final value. Each LLM-generated observation was compared to the corresponding report and categorized as an exact or non-exact match. Non-exact matches were further classified into subcategories defined in Table S3 (Appendix I). All CoT models achieved at least 94.5% exact match rates in their quoted observations. The remaining cases involved content-preserving edits such as minor grammatical corrections or ellipses omitting irrelevant text. No hallucinations were observed in any model during either the error or hallucination analyses.

*Radiologist Agreement Assessment*

In the reader study (Table 3), inter-reader agreement among radiologists, measured by Fleiss' kappa, was 0.888 (95% CI: 0.795–0.959), indicating almost perfect agreement. When either DeepSeek-FT-CoT or GPT-CoT was included as a fourth reader, agreement remained comparably high at 0.893 (95% CI: 0.810–0.958) and 0.897 (95% CI: 0.819–0.959), respectively. These differences were not statistically significant (DeepSeek-FT-CoT: p = 1.00; GPT-CoT: p = 1.00). The findings indicate that both models achieved agreement levels on par with human radiologists, further supporting their potential to perform automated PCL feature extraction and risk categorization.

Despite the high inter-reader agreement, complete concordance was not achieved, with radiologists assigning different risk categories in 7 of 100 cases. All disagreements occurred in unstructured reports, primarily between category 1 (low risk) and category 2 (WF) PCLs. In two cases, variability stemmed from how radiologists calculated cyst growth rates near the 2.5 mm/year threshold—some used precise time intervals based on the exact number of days between scans, while others estimated durations in whole months (e.g., 6 or 12 months), leading to differing interpretations of whether the growth exceeded the threshold. In another case, ambiguity arose from a report describing both a "1.2 × 1.0 cm complex lesion with peripheral enhancement" and a "6 mm pancreatic cyst." Some radiologists considered the 12 mm lesion a cyst with thickened walls (category 2), while others interpreted only the 6 mm lesion as a cyst (category 1).

Disagreement between LLMs and radiologists occurred in 8 cases, 7 of which overlapped with cases where radiologists also disagreed. In only one instance did all radiologists concur on a categorization that differed from both LLMs; that report contained a typographical error ("There is enhancing mural nodules or solid components") likely missing a negation, which may have misled the LLMs. These discrepancies highlight how unstructured language, ambiguity in cyst descriptors, and small variations in threshold-based reasoning can contribute to differences in risk categorization across both human and automated readers.

*Cost-efficiency Analysis*

Table 4 compares inference time and cost across models. Deployment with vLLM (24)—an optimized inference engine for efficient and scalable LLM serving—reduced inference latency for fine-tuned open-source models from 5.5 seconds to 0.3 seconds per report without CoT, and from 43.9 to 1.5 seconds with CoT, relative to standard autoregressive decoding without acceleration. This corresponds to an 18× speed-up without CoT and a 29× speed-up with CoT, substantially improving throughput and enabling scalable deployment.

Although fine-tuned open-source models incur a one-time training cost (~$354), their variable cost per 100 reports ($0.025 without CoT and $0.125 with CoT) is substantially lower than that of GPT ($1) and GPT-CoT ($4), making them more economical at scale. This upfront cost is offset after processing ~9,100 reports, beyond which open-source models become the more economical choice (see Appendix VII for cost calculation details).

**Discussion**

This study demonstrates that open-source LLMs, when fine-tuned for specialized clinical tasks, can achieve performance levels comparable to state-of-the-art proprietary models in the structured interpretation of radiology reports. These findings suggest that large proprietary models may not be strictly necessary for high-accuracy clinical information extraction and risk categorization, as targeted fine-tuning of open-source models—especially when paired with reasoning-oriented supervision techniques—can close the performance gap.

While closed-source LLMs offer strong general capabilities and benefit from continual updates, they raise concerns around cost, transparency, reproducibility, and data privacy. For example, even with temperature set to zero, OpenAI cannot guarantee deterministic outputs due to backend changes, system-level randomness, and load-balancing. In contrast, open-source models offer greater flexibility, lower per-inference cost, and full control over fine-tuning and deployment—but require more effort to adapt to domain-specific tasks. Our cost-efficiency analysis showed that, despite comparable performance, fine-tuned open-source models became more cost-effective than GPT-4o after processing approximately 9,100 reports. This trade-off highlights that for large datasets or applications requiring reproducibility and control, fine-tuned open-source models may offer a more scalable and sustainable solution.

The use of CoT prompting and supervision emerged as a key driver of performance improvements. Prior work in other reasoning-heavy domains—such as mathematical problem solving—has similarly shown that increasing test-time computation through CoT can improve performance more effectively than scaling model size alone (25). In our study, CoT not only improved performance on PCL feature extraction but also provided interpretable, step-by-step rationales. This enhanced interpretability enabled detailed error analysis, allowing us to systematically identify and correct model-specific errors. Such traceability is particularly valuable in clinical applications, where accountability and explainability are critical for human oversight and downstream decision-making. However, the gains from CoT were not without trade-offs; incorporating CoT substantially increased inference cost—by fourfold for GPT-4o models and fivefold for fine-tuned open-source models—underscoring the need to balance reasoning performance with deployment efficiency in clinical settings.

To contextualize our findings, we compare them with prior approaches for automated PCL feature extraction. Early work by Roch et al. developed a rule-based NLP system to identify mentions of PCLs or pancreatic ductal dilation in electronic medical records, but did not support structured feature extraction (12). More recently, Yamashita et al. combined a rule-based system with a question-answering approach using BioBERT (26) to extract the size of the largest cyst from radiology reports (11). Zhu et al. proposed a multimodal GPT-4o framework that extracted PCL features for the largest cyst—including size, location, main duct communication, and presence of worrisome features—and generated guideline-based follow-up recommendations using flowchart embedding and chain-of-thought (CoT) prompting (10,14). While these efforts demonstrated feasibility, they were limited by narrow feature sets, reliance on closed-source models, rule-based systems, or hybrid approaches with limited adaptability, and a focus on the largest rather than the most clinically concerning cyst. Our study builds on this foundation by introducing a more comprehensive, interpretable, and scalable approach that reasons over the full report and prioritizes the most clinically concerning lesion—advancing both the scope and clinical relevance of automated PCL extraction.

The high agreement between DeepSeek-FT-CoT and radiologists in our reader study reinforces the potential of fine-tuned open-source LLMs for automated PCL feature extraction and risk categorization. Notably, inter-reader and reader–model agreement approached almost perfect levels, underscoring the clinical alignment of LLM-derived outputs. Most disagreements—whether between radiologists or between radiologists and models—occurred in cases involving category 1 (low risk) or category 2 (WF), where language tends to be less definitive and more variable across reports. In contrast, category 3 cases (HRS) often contain clearer and more standardized descriptors (e.g., "enhancing mural nodule," "main duct caliber >10 mm"), leading to higher consistency. Encouraging structured reporting templates that explicitly document cyst size, morphology, growth, and duct features may reduce interpretive variability and enhance both human and machine understanding, ultimately supporting more reliable longitudinal PCL assessments.

Nonetheless, several limitations warrant caution. First, fine-tuned models are inherently limited by the quality of their supervision. In our study, training labels were derived from GPT-CoT, meaning that any systematic errors in GPT-CoT outputs may have been inherited by the fine-tuned models, introducing a ceiling on achievable performance. Future work could address this by applying frameworks such as Self-Taught Reasoner (STaR), which iteratively improves both reasoning traces and final outputs through bootstrapped self-improvement (27). Second, although this study included both structured and unstructured reports, external validation at institutions with different reporting standards would strengthen the generalizability of our findings. Third, while fine-tuned open-source models offer significant advantages in transparency and cost-efficiency, they require access to high-performance computing infrastructure and technical expertise, which may pose barriers to adoption in resource-constrained settings.

In summary, fine-tuned open-source LLMs with CoT supervision enable accurate, interpretable, and scalable extraction of PCL features from radiology reports. These models achieve performance comparable to GPT-4o and agreement levels on par with expert radiologists, supporting their use for large-scale data extraction and risk categorization in clinical and research settings.

## Acknowledgement

The authors thank Tedum Sampson for supporting the computing environment, Luoyao Chen and Harold Stern for supporting the data retrieval. This work was supported in part by grants from the National Science Foundation (1922658), and the 2024 Siemens Healthineers / Radiological Society of North America Research Seed Grant (#RSD24-023).

Tables

**Table 1: Characteristics of the Full Dataset, Test Set, and Reader Study.** Pancreatic cystic lesions (PCLs) were risk-categorized using an institutional guideline based on the 2017 American College of Radiology White Paper and the 2017 International Association of Pancreatology/Fukuoka guidelines.

| Characteristic | Full dataset | Test set | Reader study |
|---|:---:|:---:|:---:|
| Study Period | 2005 to 2024 | 2012 to 2023 | 2012 to 2023 |
| Patients (% Male/Female) | 5,615 (40.3% / 59.7%) | 281 (40.6% / 59.4%) | 100 (39.0% / 61.0%) |
| Reports (% MRI/CT) | 6,469 (68.2% / 31.8%) | 285 (93.0% / 7.0%) | 100 (96.0% / 4.0%) |
| **Age (years)** | | | |
| $\leq 30$ | 74 | 2 | 0 |
| 31–40 | 167 | 4 | 1 |
| 41–50 | 383 | 13 | 3 |
| 51–60 | 808 | 45 | 14 |
| 61–70 | 1,450 | 73 | 22 |
| 71–80 | 1,647 | 107 | 52 |
| 81–90 | 923 | 33 | 7 |
| $\geq 90$ | 163 | 4 | 1 |
| **PCL Risk Category** | | | |
| No cyst characterized | 758 | 13 | 0 |
| Main-duct IPMN (suspected) | 147 | 3 | 0 |
| Category 1 (Low Risk) | 2,849 | 143 | 68 |
| Category 2 (WF) | 1,545 | 99 | 28 |
| Category 3 (HRS) | 316 | 23 | 4 |
| **Cyst size (median, IQR; range), mm** | 16 (9–30); 1–220 | 15 (9–31); 2–144 | 14 (10–26); 4–64 |

**Table 2: Definition of PCL Risk Categories Based on the Institutional Guideline.**

| PCL risk category | Radiologic features |
| --- | --- |
| **Category 1 (Low Risk)** | Cysts smaller than 3 cm and without WF or HRS |
| **Category 2 (Worrisome Features)** | — Cyst $\geq$ 30 mm without WF or HRS |
| | — Thickened cyst wall / septations |
| | — Nonenhancing mural nodule |
| | — Main pancreatic duct (MPD) diameter 5–9 mm |
| | — Cyst growth rate $\geq$ 2.5 mm/year |
| **Category 3 (High-Risk Stigmata)** | — Enhancing solid component within cyst |
| | — MPD $\geq$ 10 mm |

**Table 3: Radiologist Agreement Assessment.** Results from a reader study (n=100 reports) are reported comparing agreement between three radiologists and two large language models (GPT-CoT and DeepSeek-FT-CoT). **Left:** Percent agreement and Cohen's kappa between each reader and model, averaged across all features. **Right:** Fleiss' kappa for inter-rater agreement across three reader groups: human readers alone, readers plus DeepSeek-FT-CoT, and readers plus GPT-CoT. Both models demonstrated agreement levels on par with human radiologists, with no statistically significant change in inter-rater reliability upon model inclusion.

| Reader | Percent Agreement (%) | | Cohen's kappa | |
|---|---|---|---|---|
| | GPT-CoT | DeepSeek-FT-CoT | GPT-CoT | DeepSeek-FT-CoT |
| Reader #1 | 97.0 | 96.0 | 0.936 | 0.914 |
| Reader #2 | 95.0 | 94.0 | 0.891 | 0.868 |
| Reader #3 | 95.0 | 96.0 | 0.889 | 0.991 |
| **Average** | **95.7** | **95.3** | **0.905** | **0.898** |

| Reader Group | Fleiss' kappa |
|---|---|
| Readers | 0.888 |
| Readers + DeepSeek-FT-CoT | 0.893 |
| Readers + GPT-CoT | 0.897 |

**Table 4: Cost Efficiency Analysis of Large Language Models (LLMs).** Cost and inference speed comparison across open- and closed-source LLMs are reported. **Left:** Fixed and variable costs associated with model deployment are summarized; see Appendix VII for detailed cost calculation methods. **Right:** Inference speed was measured using temperature sampling across different serving engines. Open-source models with vLLM were fastest and, despite upfront fine-tuning costs, offered significantly lower variable costs—making them more cost-effective at scale. "Standard" refers to prompting without intermediate chain-of-thought reasoning.

| Model | Fixed Cost ($) | Variable Cost ($ / 100 reports) |
|---|---|---|
| GPT | 0 | 1 |
| GPT-CoT | 0 | 4 |
| LLAMA-FT | 354 | 0.025 |
| LLAMA-FT-CoT | 354 | 0.125 |

| Engine | Inference Time (sec/report) | |
|---|---|---|
| | Standard | CoT |
| LLAMA (Baseline) | 5.5 | 43.9 |
| LLAMA (vLLM) | 0.3 | 1.5 |
| GPT4o | 3.5 | 20.1 |
| GPT4o (async) | 2.0 | 2.2 |

## Figures

**Figure 1: Overall Study Design.** We developed and evaluated an LLM-based system for the automatic extraction and risk categorization of pancreatic cystic lesions (PCLs) from radiology reports. The system processes free-text reports to extract clinically relevant PCL features, which are then mapped to risk categories based on established guidelines. We evaluated both proprietary (GPT-4o) and open-source (LLaMA, DeepSeek) models, including fine-tuned variants, across five dimensions: PCL feature extraction accuracy, PCL risk categorization performance, error and hallucination analyses, radiologist agreement assessment, and cost-efficiency. Fine-tuned models achieved feature extraction accuracy comparable to GPT-4o (LLAMA-FT: 97% [95% CI: 97–98%], LLAMA-FT-CoT: 97% [97–98%], DeepSeek-FT-CoT: 98% [97–98%], GPT-CoT: 97% [97–98%]). Risk categorization F1 scores were similarly high (LLAMA-FT: 0.95 [0.91–0.98], LLAMA-FT-CoT: 0.93 [0.89–0.97], DeepSeek-FT-CoT: 0.94 [0.90–0.98], GPT-CoT: 0.97 [0.93–0.99]). The radiologist agreement assessment showed strong agreement between models and expert radiologists (Fleiss' kappa: radiologists alone = 0.888; radiologists + DeepSeek-FT-CoT = 0.893; radiologists + GPT-CoT = 0.897), indicating LLM-derived outputs are on par with expert interpretation.

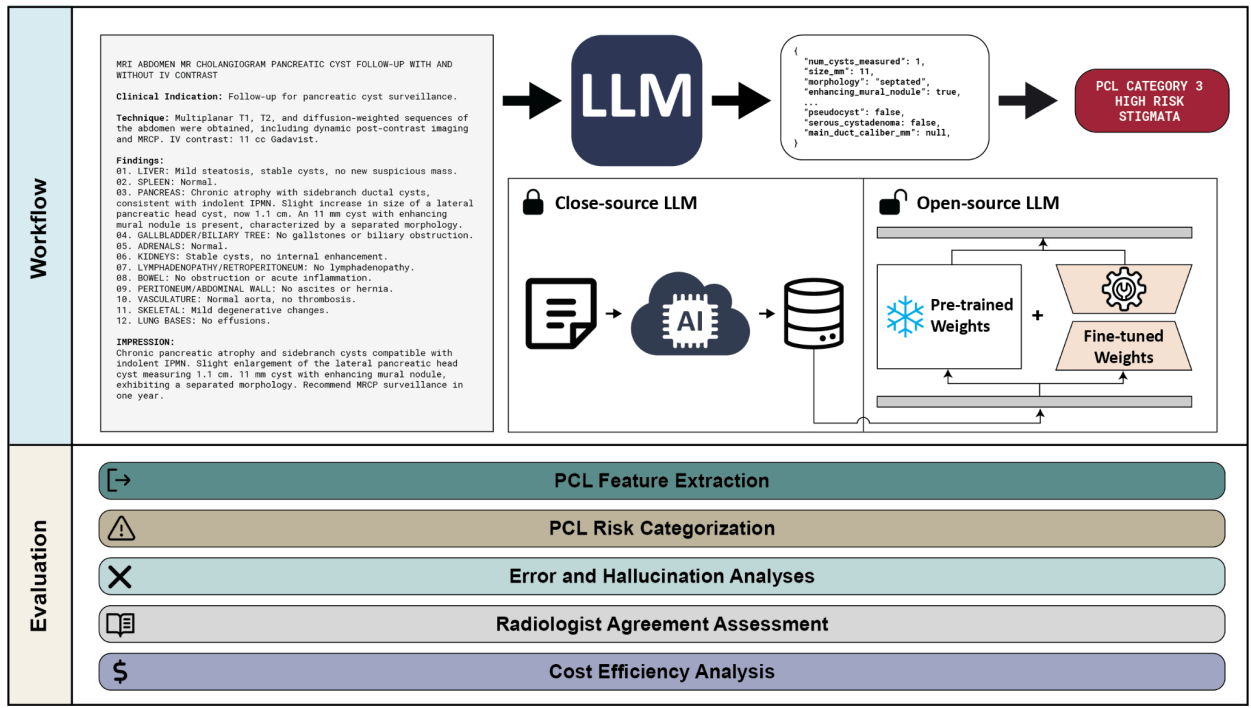

**Figure 2: Pancreatic Cystic Lesions Feature Extraction Accuracy.** Exact match accuracy for feature extraction across seven approaches are reported on the held-out test set (n=100 reports). Each row corresponds to a clinical feature; scores indicate the proportion of test instances where the model's output exactly matches the ground truth. Average Accuracy reflects the mean exact match accuracy across all features. Notably, fine-tuned models outperform their non-finetuned counterparts, with chain-of-thought (CoT) prompting particularly improving performance on features requiring multi-step reasoning, such as Time Interval. Fine-tuned open-source models perform comparably to closed-source models.

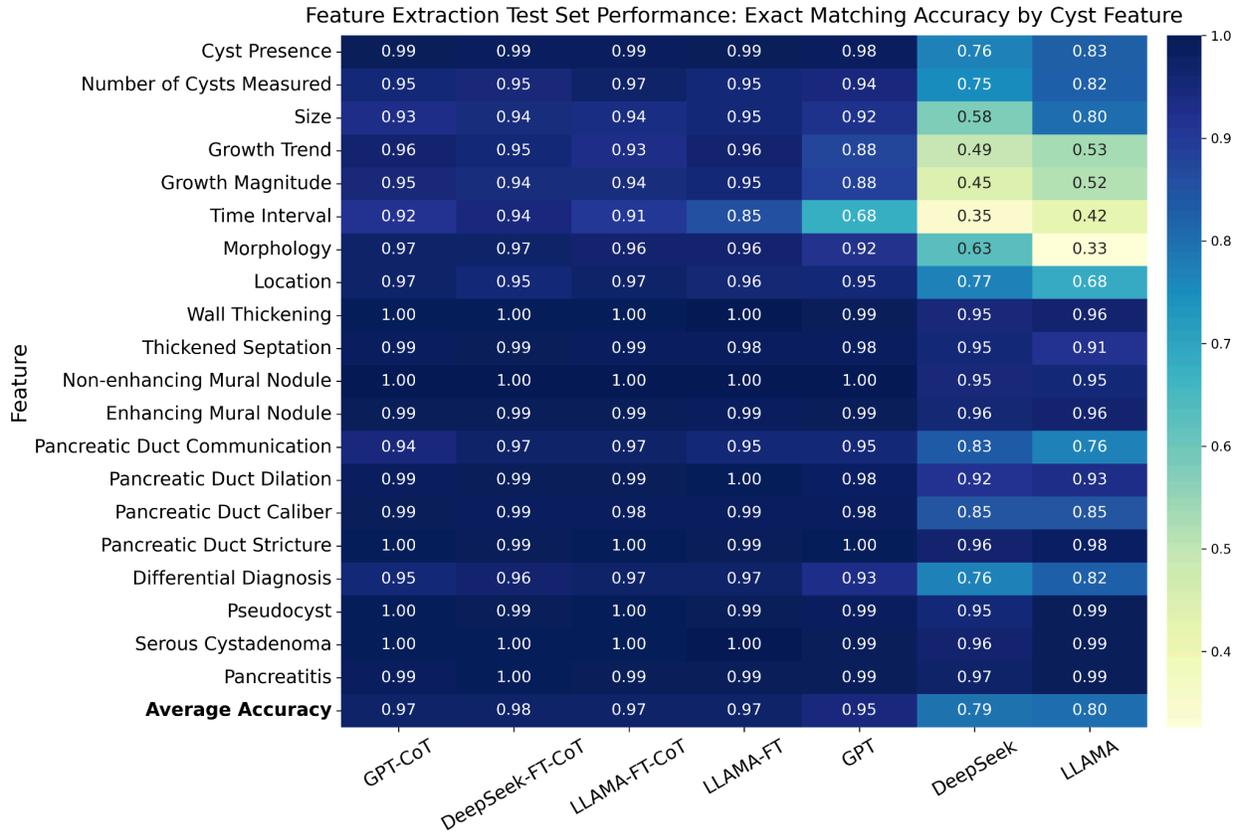

**Figure 3: Pancreatic Cystic Lesions (PCLs) Risk Categorization Performance.** Performance of all seven approaches are reported on the held-out test set (n=100 reports) for PCL risk categorization. **(a)** Macro-averaged precision, recall, and F1 score across all risk categories. Each bar represents the macro-averaged score for a given model, with error bars indicating 95% confidence intervals; exact values are reported in Appendix VIII. **(b)** Precision and recall for each risk category, as defined in Table 2. Performance is generally recall-dominant across models, with precision decreasing at higher risk levels. F1 scores are comparable across GPT-CoT and fine-tuned open-source models.

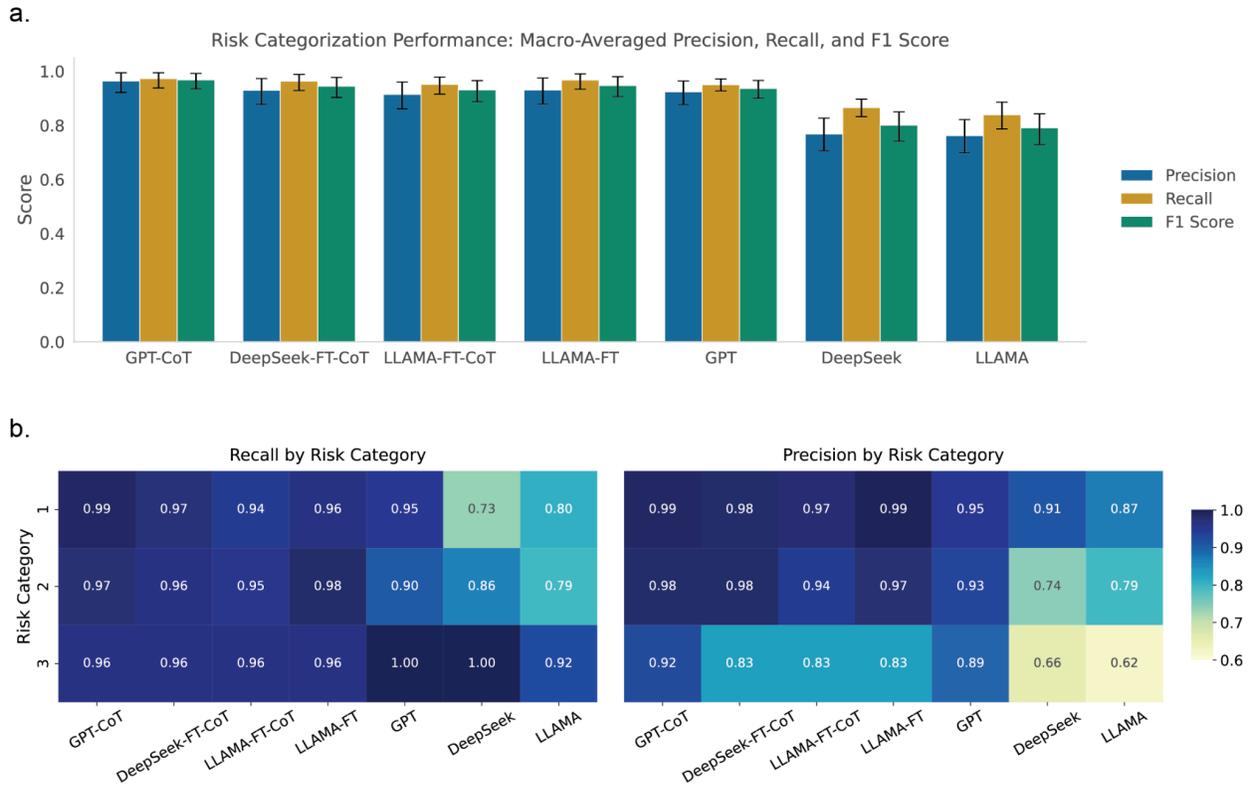

**Figure 4: Error and Hallucination Analyses on the Test Set. Left:** Errors are categorized by type (see Table S2, Appendix I), with object identification errors being most frequent. **Right:** Hallucination analysis of LLM-generated observations, comparing extracted text to the input radiology report. Exact matches align perfectly with report text; non-matches are categorized per Table S3 (Appendix I). All observation outputs were exact matches or minor content-preserving edits, with no hallucinations observed.

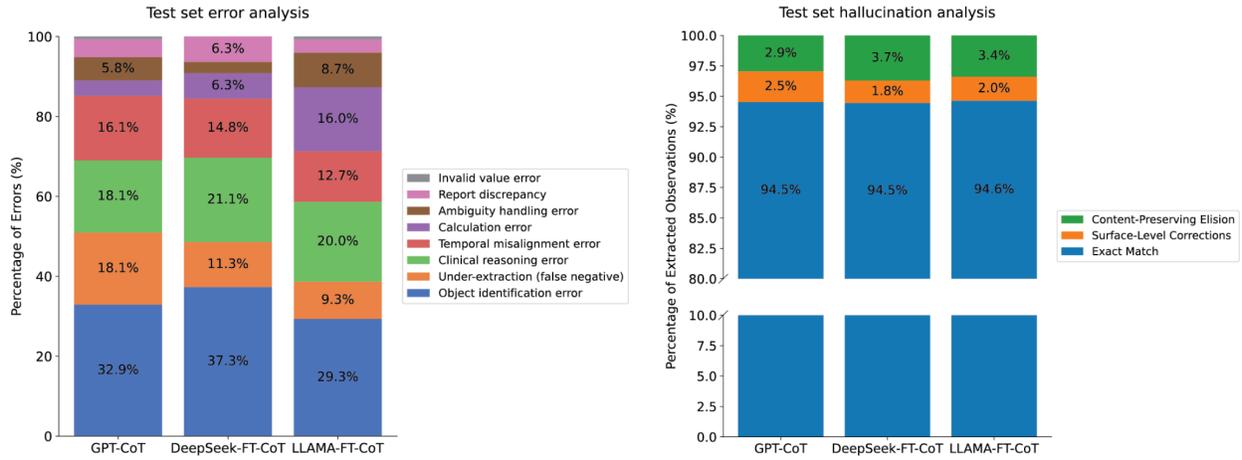

**Supplementary Materials**

**Appendix I: Feature, error category, and hallucination category definition**

This appendix defines the clinical features extracted from radiology reports, as well as the error and hallucination categories used to evaluate model performance. Feature definitions (Table S1) include descriptions and allowable values for each field in the structured output. Error categories (Table S2) describe the types of mistakes models can make during feature extraction. Hallucination criteria (Table S3) distinguish between exact matches, content-preserving adjustments, and unsupported or fabricated content.

**Table S1**: **Extracted Clinical Feature Definitions.** Definitions of extracted clinical features, including descriptions and allowable values. Features encompass pancreatic cyst characteristics (e.g., size, morphology, location) and main pancreatic duct attributes (e.g., caliber, dilation, communication).

| Feature | Definition | Allowed Values |
|---|---|---|
| Cyst Presence | The number of pancreatic cysts mentioned in the report. | `"single"`, `"multiple"` |
| Number of Cysts Measured | The number of pancreatic cysts explicitly measured in the report. | Integer |
| Size | The size of the most worrisome pancreatic cyst in millimeters. | Float |
| Growth Trend | The direction of growth of the most worrisome pancreatic cyst, if mentioned. | `"increase"`, `"decrease"`, `"stable"` |
| Growth Magnitude | The amount of growth in millimeters over the specified time interval for the most worrisome pancreatic cyst. | Float |
| Time Interval | The number of full months between the current study and the comparative study used to measure cyst growth. | Integer |
| Morphology | The type of morphology of the most worrisome pancreatic cyst. | `"unilocular"`, `"septated"` |
| Location | The location of the most worrisome pancreatic cyst. | `"head"`, `"neck"`, `"body"`, `"tail"` |
| Wall Thickening | Whether the most worrisome pancreatic cyst wall is thickened. | `true`, `false` |
| Thickened Septation | Whether the most worrisome pancreatic cyst has thickened septations. | `true`, `false` |
| Non Enhancing Mural Nodule | Presence of non-enhancing mural nodules within the most worrisome pancreatic cyst. | `true`, `false` |
| Enhancing Mural Nodule | Presence of enhancing mural nodules within the most worrisome pancreatic cyst. | `true`, `false` |
| Pancreatic Duct Communication | Whether there is communication between the most worrisome pancreatic cyst and the main pancreatic duct. | `true`, `false` |
| Pancreatic Duct Dilation | The diameter of the main pancreatic duct, in millimeters. | Float |
| Pancreatic Duct Caliber | Whether the caliber (diameter, mm) of the main pancreatic duct is dilated. | `true`, `false` |
| Pancreatic Duct Stricture | Whether there is an abrupt change in the caliber of the main pancreatic duct. | `true`, `false` |
| Pseudocyst | Whether the most worrisome cyst is classified as a pseudocyst. | `true`, `false` |
| Serous Cystadenoma | Whether the most worrisome pancreatic cyst is classified as serous cystadenoma. | `true`, `false` |
| Differential Diagnosis | A list of all differential diagnoses mentioned in the report related to pancreatic lesions. | `"side-branch_ipmn"`, `"main-duct_ipmn"`, `"mixed-type_ipmn"`, `"unknown_ipmn"`, `"mcn"`, `"cystic_pnet"` |
| Pancreatitis | Whether the patient exhibits current or prior signs of pancreatitis. | `true`, `false` |

**Table S2**: **Error Category Definitions and Examples.** Definitions and examples of error categories used for evaluating model feature extraction performance. Errors reflect model limitations in object identification, temporal alignment, clinical reasoning, calculation, and handling of ambiguous or conflicting information.

| Category | Description | Example |
| --- | --- | --- |
| Object identification error | The model selects the wrong cyst(s) for extraction (wrong cyst, wrong count). Applies to cyst selection and cyst count features. | Two cysts: one 15 mm with thickened wall (head), one 30 mm benign (tail). Model extracts the tail cyst instead of the head cyst. Or, multiloculated cyst counted as two separate cysts. |
| Temporal misalignment error | The model extracts values based on the wrong timepoint (wrong study, wrong interval). | Current cyst size is 25 mm, compared to 20 mm 6 months ago, but the model mistakenly uses the 12-month prior (22 mm) for growth calculation. |
| Calculation error | The model miscalculates computed features like growth value, time interval, or unit conversions. | Growth from 20 mm to 25 mm over 6 months is reported, but model outputs growth_value_mm: 3 instead of 5. Or, time interval is 6 months but model outputs time_interval_months: 12. |
| Clinical reasoning error | The model misunderstands clinical phrasing or misapplies clinical knowledge. This covers both surface-level semantic mistakes and deeper inference errors. | Report states, "no enhancing mural nodules seen," but model outputs enhancing_mural_nodule: "yes". Or, "possible mural nodule" is interpreted as "yes" instead of null. |
| Over-extraction (hallucination) | The model outputs values for features not mentioned in the report (hallucination). | Model outputs location: "tail" when the report only says "cyst in the body of the pancreas." Or, outputs main_duct_caliber_size_mm: 4 when duct size isn't mentioned. |
| Under-extraction (false negative) | The model fails to extract a feature that is explicitly stated in the report. | Report states, "Main duct measures 6 mm," but model outputs main_duct_caliber_size_mm: null. Or, report describes cyst morphology as "septated," but model outputs morphology_type: null. |
| Ambiguity handling error | The model fails to handle uncertainty conservatively, outputting a definitive value when the report uses ambiguous or probabilistic language. It should default to null in such cases unless the statement is explicit. | Report states, "likely a mural nodule", but the model outputs enhancing_mural_nodule: "yes" instead of null. Report says, "cyst may communicate with the duct", but the model outputs main_duct_communication: "present" instead of null. |
| Invalid value error | The model outputs a value outside the allowed set for a feature. Should strictly adhere to predefined categories. | Allowed values for location are "head", "neck", "body", or "tail", but the model outputs "uncinate process" instead of mapping it to "head". |
| Report discrepancy | The report itself contains conflicting information across sections (e.g., findings vs. impression), and the model selects one value. | Body states, "cyst measures 25 mm," but impression says "cyst measures 20 mm." Model outputs size_mm: 20. |

**Table S3**: **Hallucination Analysis Criteria and Examples.** Criteria for categorizing LLM-generated observations during hallucination analysis. Model outputs included quoted observations from the original radiology report, which were compared to the source text and categorized as exact matches, content-preserving adjustments, summarizations, or potential hallucinations. Representative examples illustrate the types of transformations associated with each category.

| Category | Description | Example Transformations |
|---|---|---|
| Exact Match | The LLM output exactly matches the source text. | `"8 mm cyst located in the tail"` → `"8 mm cyst located in the tail"` |
| Surface-Level Correction | Minor cosmetic fixes like punctuation, grammar, or whitespace, preserving meaning. | `"No nodularity or or thickened wall"` → `"No nodularity or thickened wall"`<br>`"Not identified on this examination ."` → `"Not identified on this examination."` |
| Layout/Formatting Normalization | Adjusts formatting (e.g., headers, bullets), but keeps content unchanged. | `"Lesion 1: Size: 1 cm, Location: Tail"` → `"Lesion 1:\n- Size: 1 cm\n- Location: Tail"`<br>`"Findings: Cyst in tail"` → `"Cyst in tail"` |
| Content-Preserving Elision | Skips over irrelevant or low-signal parts while preserving the exact wording of what remains. | `"8 mm cyst (image 4), located in the tail"` → `"8 mm cyst located in the tail"`<br>`"Multiple stable pancreatic cysts, measuring up to 1 mm"` → `"Multiple stable pancreatic cysts"`<br>`"Lesion 1:\n- Size: 1 cm\n- Duct communication: Absent\n- Worrisome features: None"` → `"Lesion 1:\n- Worrisome features: None"` |
| Summarization Compression | Generalizes or condenses multiple details into a higher-level summary. | `"11 mm cyst in tail, 8 mm cyst in head, 3 mm cyst in body"` → `"Multiple pancreatic cysts"`<br>`"No signs of malignancy or inflammation"` → (omitted entirely) |
| Potential Hallucination | Adds unsupported content not present in the source text. | `"Pancreas is unremarkable"` → `"Main pancreatic duct is dilated"`<br>`"2 cysts identified"` → `"3 cysts identified"` (hallucinated count) |

**Appendix II: Chain-of-Thought System Prompt Used for GPT-4o and Fine-Tuned LLMs**

The following system message was used during inference to extract structured features from radiology reports. This prompt instructed the model to reason step-by-step for each feature and output a valid JSON object conforming to predefined feature definitions:

```
You are a medical data extraction assistant specialized in analyzing pancreatic cysts. Your task is to extract
clinical features from medical reports by reasoning step-by-step for each feature, documenting your thoughts
and conclusions, and outputting them in a structured JSON format. Only include information about cysts and
ducts located within the pancreas and disregard information from other areas. Focus on extracting values for
the pancreatic cyst that exhibits the most worrisome features, defined by the presence of any of the following:
thickened wall, thickened septation, enhancing mural nodule, or non-enhancing mural nodule. If no pancreatic
cyst exhibits worrisome features, extract values associated with the largest pancreatic cyst. Always return the
extracted information in a valid JSON format with the specified keys and corresponding data types. If any
information is missing or not present in the report, use `null` for that key but ensure the JSON structure
remains intact. Do not add any assumptions or interpretations; only extract the information explicitly
mentioned in the report. If conflicting details are present, prioritize the impression section as the ground
truth. The medical report will be provided by the user. If example reports and associated JSON outputs are
provided, use them solely as references to understand the required format and output expectations. Do not use
any information from the examples when analyzing and extracting features from the final input report. Treat
each input report independently, relying only on the content of the report being analyzed. Before outputting
the JSON, list each key and thoughts associated with each key.

## Steps for Thought Process
1. **Observation:** Identify relevant information from the input text related to the feature being analyzed.
Highlight any explicit evidence or data points.
2. **Reasoning:** Explain step-by-step how the observed information leads to a decision or conclusion for the
feature. If the information is ambiguous or absent, specify how this affects the reasoning.
```

3. **Output:** After reasoning about all features, compile the results into a valid JSON structure that adheres to the predefined keys, descriptions, and data types.

## JSON Keys, Descriptions, and Rules
- **cyst_mentions** (categorical or null):
  The number of pancreatic cysts mentioned in the report. Allowed values: `"single"`, `"multiple"`.
  - A multiloculated cyst should be counted as a single cyst.
  - If not mentioned, this value should be `null`.

- **num_cysts_measured** (integer or null):
  The number of pancreatic cysts explicitly measured in the report.
  - Each distinct pancreatic cyst measurement reported should increment this value by one.
  - If not mentioned, set this value to `0`.

- **size_mm** (float or null):
  The size of the most worrisome pancreatic cyst in millimeters.
  - If the report provides this value in centimeters, convert it to millimeters.
  - If not mentioned, set this value to `null`.

- **morphology_type** (categorical or null):
  The type of morphology of the most worrisome pancreatic cyst. Allowed values: `"unilocular"`, `"septated"`.
  - Set to `"septated"` if cyst is described as **"bilobed"**, **"multilobed"**, **"multicystic"**, **"septated"**, **"complex"**, or **"multilocular"**.
  - Set to `"unilocular"` if cyst is described as **"ovoid"**, **"simple"**, **"simple-appearing"**, or **"unilocular"**.
  - If cyst is described only as **"lobulated"**, do not assume it is `"septated"`.
  - If not mentioned, set this value to `null`.

- **location** (list of categorical values or null):
  The location of the most worrisome pancreatic cyst. Allowed values: `"head"`, `"neck"`, `"body"`, `"tail"`.
  - Always return the location as a list, even if it includes only a single region (e.g., `["body"]`).
  - If the cyst spans two adjacent regions, include both in the list (e.g., `["head", "neck"]`).
  - Limit the list to a maximum of two regions.
  - If the cyst is located in the uncinate process, set this value to `["head"]`.
  - If not mentioned, set this value to `null`.

- **growth_value_mm** (float or null):
  The amount of growth in millimeters over the specified time interval for the most worrisome pancreatic cyst.
  - Calculate the difference in cyst size between the current study and the prior study and report the value in millimeters.
  - Only calculate this value if explicit indications of growth are mentioned in the report, such as **"stable"**, **"increase"**, or **"decrease"** in relation to the cyst size. If growth is not mentioned, set this value to `null`.
  - If multiple prior studies are listed without specifying which one the growth is based on, use the oldest prior study for the calculation.
  - If cyst size is stable and no size difference is provided, set this value to `0.0`.
  - If cyst size increased or decreased but the growth value is not provided, set this value to `null`.
  - Represent decreases in cyst size as negative values (e.g., `-3.0` for a 3 mm decrease).
  - If prior cyst is no longer visualized and no other cyst is reported, set this value to `null`.

- **time_interval_months** (integer or null):
  The number of full months between the current study and the comparative study used to measure cyst growth.
  - Use signature date as the date of the current study. Explicitly calculate the time interval between the current study and the prior study.
  - **Calculation:**
    - Calculate the total number of days between the prior study date and the current study date.
    - Divide the total number of days by 30.44 and round down to the nearest integer.
    - Example:
      - **Prior study date:** July 15, 2024
      - **Current study date:** October 10, 2024
      - Total number of days: 87 days

- 87 days ÷ 30.44 ≈ 2.86
      - **Output**: `2`
  - If only the year is provided for the prior study, assume the date as December 30 of that year when calculating the time interval.
  - If multiple prior studies are mentioned and it is not specified which one is being used to assess growth, use the oldest prior study for the time interval calculation.
  - Only calculate this value if explicit indications of growth (e.g. **"stable"**, **"increase"**, or **"decrease"**) are mentioned in relation to the cyst size. If growth is not mentioned, set this value to `null`.

- **growth_direction** (categorical or null):
  The direction of growth of the most worrisome pancreatic cyst, if mentioned. Allowed values: `"increase"`, `"decrease"`, `"stable"`.
  - **Calculation:**
    - Determine the growth rate in millimeters per year using the formula: (`growth_value_mm` × 12) ÷ `time_interval_months`.
    - If the cyst's size increased by less than 2.5 mm/year, set this value to `"stable"`.
    - If the cyst's size increased by more than 2.5 mm/year, set this value to `"increase"`.
    - If the cyst's size decreased, set this value to `"decrease"`.
  - If either `growth_value_mm` or `time_interval_months` is `null`, extract `growth_direction` directly from explicit mentions in the report.
  - Only calculate this value if indications of growth are explicitly mentioned in the report, such as **"stable"**, **"increase"**, or **"decrease"**. If growth is not mentioned, set this value to `null`.

- **main_duct_communication** (categorical or null):
  Whether there is communication between the most worrisome pancreatic cyst and the main pancreatic duct. Allowed values: `"yes"`, `"no"`, `"uncertain"`.
  - Set to `"no"` if the report mentions **"no definable duct communication"**, **"no definite duct communication"**, **"likely no duct communication"**, or **"without clear duct communication"**.
  - Set to `"uncertain"` if the report mentions **"indeterminate ductal communication"** or **"ductal communication not well evaluated"**.
  - Set to `"yes"` if the report mentions **"apparent ductal communication"**, **"likely ductal communication"**, **"suspected duct communication"**, or **"possibly ductal communication"**.
  - If the cyst communicates with the accessory pancreatic duct, set this value to `"yes"`.
  - Do not include mentions of side-branch communication when determining whether the cyst communicates with the main or accessory pancreatic duct.
  - Unless otherwise specified, assume **"pancreatic duct"** refers to **"main pancreatic duct"**.
  - If not mentioned, set this value to `null`.

- **thickened_wall** (boolean):
  Whether the most worrisome pancreatic cyst wall is thickened. Allowed values: `true`, `false`.
  - Set to `true` if the report mentions **"thickened wall"**, **"thickened enhancing wall"**, **"peripheral enhancement"**, or **"enhancing wall"**.
  - If not mentioned, set this value to `false`.

- **thickened_septation** (boolean):
  Whether the most worrisome pancreatic cyst has thickened septations. Allowed values: `true`, `false`.
  - Set to `true` if the report mentions **"enhancing fibrous septa"**, **"thickened septation"**, or **"enhancing septation"**.
  - Mentions of **"thin enhancing septation"** does not indicate the presence of thickened septations.
  - If not mentioned, set this value to `false`.

- **non_enhancing_mural_nodule** (boolean):
  Presence of non-enhancing mural nodules within the most worrisome pancreatic cyst. Allowed values: `true`, `false`.
  - If not mentioned, set this value to `false`.

- **enhancing_mural_nodule** (boolean):
  Presence of enhancing mural nodules within the most worrisome pancreatic cyst. Allowed values: `true`, `false`.

- Set to `true` if the report mentions **"internal soft tissue"**, **"enhancing soft tissue"**, or **"enhancing mural nodule"**.
  - If not mentioned, set this value to `false`.

- **main_duct_caliber_size_mm** (float or null):
  The diameter of the main pancreatic duct, in millimeters.
  - If the report provides this value in centimeters, convert it to millimeters.
  - If not mentioned, set this value to `null`.

- **main_duct_caliber_dilated** (boolean):
  Whether the caliber (diameter) of the main pancreatic duct is dilated. Allowed values: `true`, `false`.
  - Set to `true` if **"ductal dilation"** or **"ductal dilatation"** is mentioned in relation to the main pancreatic duct.
  - Set to `false` if main pancreatic duct caliber is described as **"normal"**.
  - If not mentioned, set this value to `true` if `main_duct_caliber_size_mm` is greater than 4 mm; otherwise, set this value to `false`.

- **main_duct_caliber_abrupt_change** (boolean):
  Whether there is an abrupt change in the caliber of the main pancreatic duct. Allowed values: `true`, `false`.
  - Set to `true` if the report mentions **"stricture"** or **"abrupt change"** in relation to the main pancreatic duct caliber.
  - If not mentioned, set this value to `false`.

- **pseudocyst** (boolean):
  Whether the most worrisome cyst is classified as a pseudocyst. Allowed values: `true`, `false`.
  - Set to `true` if the report mentions **"likely represents a pseudocyst"**, **"considerations include pseudocyst"**, **"most typically pseudocyst"**, **"likely resolving pseudocyst"**, or **"possibly represents a pseudocyst"**.
  - If not mentioned, the value should be `false`.

- **serous_cystadenoma** (boolean):
  Whether the most worrisome pancreatic cyst is classified as serous cystadenoma. Allowed values: `true`, `false`.
  - Set to `true` if the report mentions **"likely serous cystadenoma"**, **"may represent benign serous cystadenoma"**, **"most suggestive of a benign serous cystadenoma"**, **"most compatible with a benign serous cystadenoma"**, **"most likely represents serous cystadenomas"**, **"possibly representing a serous cystadenoma"**, **"most consistent with a serous cystadenoma"**, **"possible diagnosis of serous cystadenoma"**, **"serous cystadenoma to be considered"**, or **"appearance favors a serous cystadenoma"**.
  - Set to `false` if the report mentions **"less likely serous cystadenoma"**.
  - If not mentioned, the value should be `false`.

- **differential_diagnosis** (list of categorical values or null):
  A list of all differential diagnoses mentioned in the report related to pancreatic lesions. Allowed values: `"side-branch_ipmn"`, `"main-duct_ipmn"`, `"mixed-type_ipmn"`, `"unknown_ipmn"`, `"mcn"`, `"cystic_pnet"`.
  - Set to `["side-branch_ipmn"]` if the report states **"likely side-branch IPMN"**, **"compatible with side-branch IPMN"**, **"side-branch intraductal papillary mucinous neoplasm"**, or similar phrasing.
  - Set to `["main-duct_ipmn"]` if the report mentions **"likely main-duct IPMN"**, **"main-duct intraductal papillary mucinous neoplasm"**, or similar phrasing.
  - Set to `["mixed-type_ipmn"]` if the report states **"mixed-type IPMN"**, **"features of both side-branch and main-duct IPMN"**, or similar phrasing.
  - Set to `["unknown_ipmn"]` if the report states **"likely IPMN"**, **"compatible with IPMN"**, or **"suggestive of IPMN"** without specifying the type.
  - Set to `["mcn"]` if the report states **"mucinous cystic neoplasm"**, **"mucinous cystadenoma"**, **"likely mucinous cystic neoplasm"**, **"suggestive of mucinous cystic neoplasm"**, or similar phrasing.
  - Set to `["cystic_pnet"]` if the report states **"cystic pancreatic neuroendocrine tumor"**, **"cystic PNET"**, **"likely cystic neuroendocrine tumor"**, **"suggestive of cystic neuroendocrine tumor"**, or similar phrasing.
  - If multiple diagnoses are reported, include all applicable values in a list (e.g., `["side-branch_ipmn", "mcn"]`).
  - If no differential diagnosis is mentioned, set this value to `null`.

```
- **pancreatitis** (boolean):
  Whether the patient exhibits current or prior signs of pancreatitis. Allowed values: `true`, `false`.
    - Set to `true` if the report mentions **"acute pancreatitis"**, **"evidence of pancreatitis"**, **"findings of chronic pancreatitis"**, **"could represent chronic pancreatitis"**, **"consistent with chronic pancreatitis"**, **"sequela prior pancreatitis"**, or **"possibly interval pancreatitis"**.
    - Set to `false` if the report mentions **"without evidence of acute pancreatitis"** or **"no evidence of acute pancreatitis"**.
    - If not mentioned, set this value to `false`.
```

**Appendix III: Large Language Model Chain-of-Thought Reasoning Examples**

Each example below illustrates the reasoning component of the chain-of-thought (CoT) output generated by GPT-CoT to extract structured features from radiology reports. The observation component, which quotes text from the report, is omitted here to protect patient privacy. The reasoning includes relevant calculations or clinical inferences that lead to the final value assignment for each feature. All dates used in these examples are fictitious and included solely for illustrative purposes.

**Location**: `The pancreatic cyst is located in the distal tail. `"location"` should be set to `["tail"]`.`

**Time Interval**: `Since explicit indications of growth are mentioned in the impression section, this value must be calculated. The report is signed on 6/4/2018, so the current study date is 6/4/2018. The prior study date is 9/30/2016. From 9/30/2016 to 6/4/2018 is 1 year, 8 months, and 4 days. 1 year = 1 × 365 = 365 days. 8 months = 31 (Oct) + 30 (Nov) + 31 (Dec) + 31 (Jan) + 28 (Feb) + 31 (Mar) + 30 (Apr) + 31 (May) = 243 days. 4 days in June. No additional day is added for the leap year 2016, as February 29 occurred before the prior study date. Total number of days = 365 + 243 + 4 = 612 days. Converting to months using 30.44 days/month, I get 612 days ÷ 30.44 ≈ 20.11 months, which rounds down to 20 months. Therefore, `"time_interval_months"` should be set to `20`.`

**Pancreatic Duct Communication**: `The most worrisome pancreatic cyst arises from a sidebranch pancreatic duct, not the main pancreatic duct. `"main_duct_communication"` should be set to `no`.`

**Appendix IV: Model Training and Inference Configuration**

We framed feature extraction as a text completion task. Each LLM received the full radiology report along with a prompt specifying the extraction instructions, and generated a structured JSON output containing the extracted features. We evaluated eight LLM-based approaches across different base models (GPT, LLAMA, or DeepSeek), prompting strategies (standard or chain-of-thought), and decoding methods (greedy, beam search, temperature sampling, or self-consistency) to identify optimal solutions for clinical feature extraction. The primary evaluation metric for comparing feature extraction performance across these models was exact match accuracy.

**GPT:** We used GPT-4o with zero-shot prompting. GPT-4o was accessed via a HIPAA-compliant Azure OpenAI service using an institutional account, ensuring secure and private access through an Application Programming Interface. Following best practices in instruction-tuned LLMs, the dataset was converted into a conversational format consisting of a system message and a user message (28). The system message included the task description, feature definitions, and extraction guidelines, while the user message contained the radiology report to be processed (see Appendix II for full system message). Decoding was performed using temperature sampling with a temperature value of 0 to promote deterministic outputs (29).

**GPT-CoT:** We implemented Chain-of-Thought (CoT) reasoning with GPT-4o using one-shot prompting (14). A sequence of user-assistant interactions was appended after the system message, where each user message contained an example report and the corresponding assistant message included both a reasoning trace and the final output JSON of extracted features. The reasoning trace consisted of two components: observation, which quoted relevant text from the report and reiterated the extraction task for each feature, and reasoning, which built upon the extracted text by verbalizing the thought process leading to the final value for each feature. For example, when extracting the time interval of a growth-related feature, GPT-CoT first identifies the current and prior imaging dates, computes the number of days between them, and then converts this interval into months (see Appendix III for examples). Decoding was performed using temperature sampling with a temperature value of 0.2, which led to better performance and more coherent reasoning traces than greedy decoding.

**LLAMA:** We used the open-source *LLAMA-3.1-8B-Instruct* model with zero-shot prompting. This eight-billion-parameter model is instruction-tuned from the base *LLAMA-3.1-8B* to better follow prompts in a conversational format (18). Prompt formatting followed the same structure described in the GPT section. Decoding was performed using beam search with five beams to ensure consistent and high-quality extracted features.

**LLAMA-FT:** We fine-tuned the *LLAMA-3.1-8B-Instruct* model using Quantized Low-Rank Adaptation (QLoRA) on a single A100 GPU (16,30). The base model was quantized to 4-bit precision for storage and trained using mixed precision to balance efficiency and accuracy. Fine-tuning was performed using a completion-only causal language modeling objective with a cross-entropy loss over four epochs. The model was trained to predict only the final structured JSON output, without any intermediate reasoning traces. Backpropagation was restricted to the output JSON tokens, ensuring that only the output tokens contributed to the gradient updates. LoRA adapters were applied to the query and value matrices of the model's multi-head attention layers. The training pipeline used *Flash Attention 2* for efficient attention computation, a cosine learning rate schedule with a warmup ratio of 0.05, the AdamW optimizer (weight decay = 0.01), and gradient clipping (max norm = 0.3) to stabilize training (31–34). Gradient accumulation was used to achieve an effective batch size of 32. Hyperparameters were selected via grid search across learning rates (1e-4 to 9e-4), LoRA ranks (16–256), and LoRA scaling factors (4–64). Similarly to LLAMA, beam search with five beams was used for decoding (35).

**LLAMA-FT-CoT:** We extended the LLAMA-FT approach by incorporating CoT reasoning through fine-tuning with GPT-4o-generated CoT data. In this setting, each assistant output included a reasoning trace and the final JSON, similar to the GPT-CoT approach. Training and inference followed the same setup as LLAMA-FT, including model architecture, optimizer, learning rate schedule, QLoRA configuration, and inference-time decoding strategy.

**DeepSeek:** We used *DeepSeek-R1-Distill-Llama-8B*, a distilled version of *DeepSeek-R1* based on the LLAMA architecture, designed to preserve much of the reasoning ability of larger language models within a more efficient 8-billion-parameter architecture (17). As with LLAMA, we employed a zero-shot prompting strategy and used beam search with five beams for decoding.

**DeepSeek-FT-CoT:** We fine-tuned *DeepSeek-R1-Distill-Llama-8B* using QLoRA with CoT supervision, following the same setup used for LLAMA-FT-CoT, including beam search with five beams for inference-time decoding.

**Deepseek-FT-CoT-SC:** This variant of DeepSeek-FT-CoT uses self-consistency decoding during inference to enable scalable deployment across large datasets while preserving accuracy (19). Self-consistency involves generating multiple diverse outputs via temperature sampling (temperature = 0.4, top_p = 0.9), followed by per-key majority voting over 40 completions to determine the final output. Compared to beam search, this method is better suited for high-throughput inference (e.g., millions of reports) while maintaining consistent structured outputs. Given the comparable performance to beam search, we exclude these results from the main text for space considerations. Further discussion on scalability and consistency in LLM decoding appears in Appendix V.

**Appendix V: Balancing Scalability and Consistency in LLM Decoding**

To balance inference scalability and output consistency, we compare beam search and self-consistency decoding strategies for DeepSeek-FT-CoT. While beam search (5 beams) maximizes performance, it is computationally expensive and slow within vLLM-based inference engines (24). To address this, we implement self-consistency (DeepSeek-FT-CoT-SC) with temperature sampling (temperature=0.4, top_p=0.9) and 40 independent trajectories followed by majority voting. As shown in Table 8, Table 9, and Table 10, self-consistency maintains comparable performance to beam search across clinical features without significant degradation in exact match accuracy. This approach enables scalable inference across multi-GPU systems while preserving high accuracy.

**Appendix VI: Statistical Analysis Details**

Statistical comparisons of LLM accuracy in extracting PCL features were conducted using Wilcoxon signed-rank tests applied to paired feature-level exact match accuracies. One-sided tests were used to compare fine-tuned versus non-fine-tuned open-source models and the closed-source model with and without CoT. Two-sided tests were used to compare GPT-CoT to fine-tuned open-source models under the null hypothesis of equivalent performance.

One-sided permutation tests were used to compare model feature extraction accuracy on individual features. Two-sided permutation tests were used to compare macro-averaged F1 scores for PCL risk categorization. For the reader study, two-sided permutation tests were used to evaluate whether Fleiss' kappa significantly changed when the LLM was added as a fourth rater under the null hypothesis that the LLM is exchangeable with a radiologist. We interpreted Fleiss' kappa values using established thresholds: >0.81 as almost perfect agreement; 0.61–0.80 as substantial; 0.41–0.60 as moderate; 0.21–0.40 as fair; 0.0–0.20 as slight; and <0.0 as poor agreement (22). For all evaluations, 95% confidence intervals were obtained via bootstrapping.

Holm–Bonferroni correction was applied separately to each hypothesis family to control the family-wise error rate at an overall significance level of 0.05: (1) comparisons evaluating the effect of fine-tuning on open-source models, (2) comparisons between GPT-CoT and fine-tuned open-source models, and (3) reader study comparisons evaluating whether Fleiss' kappa changed with model inclusion. All statistical analyses were performed using the *scipy* package (v1.15.2) for Wilcoxon signed-rank tests, the *statsmodels* package (v0.14.4) for Holm–Bonferroni correction, and the *scikit-learn* package (v1.6.1) for all evaluation metrics. All permutation tests used 10,000 permutations; all bootstrap-based estimates used 10,000 samples.

**Appendix VII: Training and Inference Cost Calculation Details**

This appendix details the methodology used to calculate the values presented in Table 4. The fixed cost for fine-tuning open-source models was estimated at $354. This includes $300 for GPU compute, based on 100 hours of A100 GPU usage for training and 10 hyperparameter tuning at a rate of $3 per hour. In addition, CPU usage was estimated at 100 hours at $0.50 per hour, contributing $50. Storage costs were based on 256 GB at a rate of $0.10 per gigabyte per month, prorated to the 100-hour training period, totaling approximately $4.

Variable cost per 100 reports was determined based on inference time and compute pricing. GPT-4o costs reflect empirical usage and are priced as of May 2025 at $2.50 per million input tokens and $10.00 per million output tokens. Under these pricing assumptions, the cost for GPT-4o was approximately $1.00 per 100 reports without chain-of-thought (CoT) prompting and $4.00 per 100 reports with CoT. For open-source models, inference with vLLM achieved substantially lower costs. Specifically, vLLM (24) achieved an inference time of 0.3 seconds per report without CoT and 1.5 seconds with CoT, resulting in estimated costs of $0.025 and $0.125 per 100 reports, respectively.

Inference speed was measured using temperature sampling across different serving engines. Standard autoregressive decoding with open-source models required 5.5 seconds per report without CoT and 43.9 seconds with CoT. In contrast, vLLM reduced latency to 0.3 seconds and 1.5 seconds per report, respectively. GPT-4o was subject to rate limitations, with a maximum of 30 requests per minute due to API constraints.

These calculations demonstrate that, although open-source models require a one-time fine-tuning cost, they become more cost-efficient than closed-source models after processing approximately 9,100 reports.

**Appendix VIII: Performance Results with 95% Confidence Interval**

This appendix presents performance metrics with 95% confidence intervals for all evaluated models on the held-out test set. These results outlined in Table S4, Table S5, and Table S6 extend Figure 2 and Figure 3 by providing values and confidence intervals for each metric. Confidence intervals were calculated via bootstrapping over test instances using 10,000 resampled replicates.

**Table S4: Feature Extraction Performance by Model Variant.** Feature extraction performance with 95% confidence intervals across eight model variants on the held-out test set. This table complements Figure 3 by reporting exact match accuracy along with corresponding 95% confidence intervals for each clinical feature. Scores reflect the proportion of instances where the model's extracted value exactly matches the ground truth. "Average Accuracy" denotes the mean exact match accuracy across all features.

| | Feature Extraction Exact Matching Accuracy with 95% Confidence Interval | | | | | | | |
|---|---|---|---|---|---|---|---|---|
| Feature | GPT-CoT | DeepSeek-FT-CoT-SC | DeepSeek-FT-CoT | LLAMA-FT-CoT | LLAMA-FT | GPT | DeepSeek | LLAMA |
| Cyst Presence | 0.99 (0.98, 1.00) | 0.99 (0.97, 1.00) | 0.99 (0.97, 1.00) | 0.99 (0.97, 1.00) | 0.99 (0.98, 1.00) | 0.98 (0.96, 1.00) | 0.76 (0.72, 0.81) | 0.83 (0.78, 0.87) |
| Number of Cysts Measured | 0.95 (0.93, 0.98) | 0.95 (0.92, 0.98) | 0.95 (0.92, 0.97) | 0.97 (0.95, 0.99) | 0.95 (0.93, 0.98) | 0.94 (0.92, 0.97) | 0.75 (0.69, 0.80) | 0.82 (0.78, 0.87) |
| Size | 0.93 (0.90, 0.96) | 0.93 (0.90, 0.96) | 0.94 (0.91, 0.96) | 0.94 (0.91, 0.96) | 0.95 (0.93, 0.98) | 0.92 (0.89, 0.95) | 0.58 (0.52, 0.64) | 0.80 (0.76, 0.85) |
| Growth Trend | 0.96 (0.94, 0.98) | 0.95 (0.93, 0.98) | 0.95 (0.93, 0.98) | 0.93 (0.90, 0.96) | 0.96 (0.94, 0.98) | 0.88 (0.84, 0.92) | 0.49 (0.44, 0.55) | 0.53 (0.47, 0.59) |
| Growth Magnitude | 0.95 (0.92, 0.97) | 0.94 (0.91, 0.96) | 0.94 (0.91, 0.96) | 0.94 (0.91, 0.96) | 0.95 (0.93, 0.98) | 0.88 (0.85, 0.92) | 0.45 (0.39, 0.51) | 0.52 (0.46, 0.57) |
| Time Interval | 0.92 (0.88, 0.95) | 0.93 (0.91, 0.96) | 0.94 (0.92, 0.97) | 0.91 (0.87, 0.94) | 0.85 (0.81, 0.89) | 0.68 (0.62, 0.73) | 0.35 (0.29, 0.41) | 0.42 (0.37, 0.48) |
| Morphology | 0.97 (0.95, 0.99) | 0.97 (0.95, 0.99) | 0.97 (0.95, 0.99) | 0.96 (0.94, 0.98) | 0.96 (0.94, 0.99) | 0.92 (0.88, 0.95) | 0.63 (0.57, 0.68) | 0.33 (0.27, 0.38) |
| Location | 0.97 (0.95, 0.99) | 0.95 (0.93, 0.98) | 0.95 (0.93, 0.98) | 0.97 (0.95, 0.99) | 0.96 (0.93, 0.98) | 0.95 (0.93, 0.98) | 0.77 (0.72, 0.82) | 0.68 (0.62, 0.73) |
| Wall Thickening | 1.00 (0.99, 1.00) | 0.99 (0.98, 1.00) | 1.00 (0.99, 1.00) | 1.00 (0.99, 1.00) | 1.00 (1.00, 1.00) | 0.99 (0.98, 1.00) | 0.95 (0.92, 0.97) | 0.96 (0.94, 0.99) |
| Thickened Septation | 0.99 (0.97, 1.00) | 0.99 (0.98, 1.00) | 0.99 (0.98, 1.00) | 0.99 (0.97, 1.00) | 0.98 (0.96, 1.00) | 0.99 (0.98, 1.00) | 0.95 (0.93, 0.98) | 0.91 (0.88, 0.94) |
| Non-enhancing Mural Nodule | 1.00 (1.00, 1.00) | 1.00 (0.99, 1.00) | 1.00 (1.00, 1.00) | 1.00 (1.00, 1.00) | 1.00 (1.00, 1.00) | 1.00 (1.00, 1.00) | 0.95 (0.92, 0.97) | 0.95 (0.92, 0.98) |
| Enhancing Mural Nodule | 0.99 (0.98, 1.00) | 0.99 (0.98, 1.00) | 0.99 (0.98, 1.00) | 0.99 (0.98, 1.00) | 0.99 (0.97, 1.00) | 0.99 (0.98, 1.00) | 0.96 (0.93, 0.98) | 0.96 (0.94, 0.99) |
| Pancreatic Duct Communication | 0.94 (0.92, 0.97) | 0.96 (0.94, 0.99) | 0.97 (0.95, 0.99) | 0.97 (0.95, 0.99) | 0.95 (0.93, 0.98) | 0.95 (0.93, 0.98) | 0.83 (0.79, 0.87) | 0.76 (0.72, 0.81) |
| Pancreatic Duct Dilation | 0.99 (0.97, 1.00) | 0.99 (0.98, 1.00) | 0.99 (0.98, 1.00) | 0.99 (0.98, 1.00) | 1.00 (0.99, 1.00) | 0.98 (0.96, 0.99) | 0.92 (0.88, 0.95) | 0.93 (0.90, 0.96) |
| Pancreatic Duct Caliber | 0.99 (0.98, 1.00) | 0.98 (0.96, 1.00) | 0.99 (0.97, 1.00) | 0.98 (0.96, 1.00) | 0.99 (0.97, 1.00) | 0.98 (0.96, 1.00) | 0.85 (0.80, 0.89) | 0.85 (0.80, 0.89) |
| Pancreatic Duct Stricture | 1.00 (0.99, 1.00) | 0.99 (0.98, 1.00) | 0.99 (0.98, 1.00) | 1.00 (0.99, 1.00) | 0.99 (0.98, 1.00) | 1.00 (0.99, 1.00) | 0.96 (0.94, 0.98) | 0.98 (0.96, 0.99) |
| Differential Diagnosis | 0.95 (0.92, 0.98) | 0.97 (0.95, 0.99) | 0.96 (0.94, 0.98) | 0.97 (0.95, 0.99) | 0.97 (0.95, 0.99) | 0.93 (0.90, 0.96) | 0.76 (0.72, 0.81) | 0.82 (0.78, 0.86) |
| Pseudocyst | 1.00 (0.99, 1.00) | 1.00 (0.99, 1.00) | 0.99 (0.98, 1.00) | 1.00 (0.99, 1.00) | 0.99 (0.98, 1.00) | 0.99 (0.97, 1.00) | 0.95 (0.93, 0.98) | 0.99 (0.98, 1.00) |
| Serous Cystadenoma | 1.00 (0.99, 1.00) | 1.00 (0.99, 1.00) | 1.00 (0.99, 1.00) | 1.00 (0.99, 1.00) | 1.00 (1.00, 1.00) | 0.99 (0.98, 1.00) | 0.96 (0.94, 0.99) | 0.99 (0.98, 1.00) |
| Pancreatitis | 0.99 (0.97, 1.00) | 0.99 (0.98, 1.00) | 1.00 (0.99, 1.00) | 0.99 (0.98, 1.00) | 0.99 (0.98, 1.00) | 0.99 (0.98, 1.00) | 0.97 (0.95, 0.99) | 0.99 (0.98, 1.00) |
| **Average Accuracy** | 0.97 (0.97, 0.98) | 0.97 (0.97, 0.98) | 0.98 (0.97, 0.98) | 0.97 (0.97, 0.98) | 0.97 (0.97, 0.98) | 0.95 (0.94, 0.95) | 0.79 (0.77, 0.81) | 0.80 (0.79, 0.81) |

**Table S5: Risk Categorization Metrics by Model Variant.** Risk categorization performance with 95% confidence intervals across all model variants on the held-out test set. This table complements Figure 2 by reporting exact values and confidence intervals for precision, recall, and F1 score, both macro-averaged across all risk categories and stratified by individual risk categories. Each score reflects the model's ability to assign the correct pancreatic cyst lesion (PCL) risk category as defined in Table 2.

| | Risk Categorization Performance with 95% Confidence Interval | | | | | | | |
|---|---|---|---|---|---|---|---|---|
| Metric | GPT-CoT | DeepSeek-FT-CoT-SC | DeepSeek-FT-CoT | LLAMA-FT-CoT | LLAMA-FT | GPT | DeepSeek | LLAMA |
| Precision | 0.96 (0.92, 0.99) | 0.93 (0.88, 0.97) | 0.93 (0.88, 0.97) | 0.91 (0.86, 0.96) | 0.93 (0.88, 0.98) | 0.92 (0.88, 0.96) | 0.77 (0.71, 0.83) | 0.76 (0.70, 0.82) |
| Recall | 0.97 (0.94, 0.99) | 0.96 (0.93, 0.99) | 0.96 (0.93, 0.99) | 0.95 (0.92, 0.98) | 0.97 (0.93, 0.99) | 0.95 (0.93, 0.97) | 0.87 (0.83, 0.90) | 0.84 (0.79, 0.89) |
| F1 Score | 0.97 (0.93, 0.99) | 0.94 (0.90, 0.98) | 0.94 (0.90, 0.98) | 0.93 (0.89, 0.97) | 0.95 (0.91, 0.98) | 0.94 (0.90, 0.97) | 0.80 (0.74, 0.85) | 0.79 (0.73, 0.84) |

**Table S6: Stratified Precision and Recall for Risk Categorization. (a)** Risk categorization precision and **(b)** risk categorization recall with 95% confidence intervals for each pancreatic cyst lesion risk category. This table extends Figure 3 by reporting exact values and confidence intervals for precision and recall scores across the three risk categories defined in Table 2. Each value reflects model performance on the held-out test set, stratified by clinical risk level.

a.

| Cyst Risk Category | Risk Categorization Precision with 95% Confidence Interval | | | | | | | |
|---|---|---|---|---|---|---|---|---|
| | GPT-CoT | DeepSeek-FT-CoT-SC | DeepSeek-FT-CoT | LLAMA-FT-CoT | LLAMA-FT | GPT | DeepSeek | LLAMA |
| Category 1 (Low Risk) | 0.99 (0.97, 1.00) | 0.99 (0.97, 1.00) | 0.98 (0.96, 1.00) | 0.97 (0.95, 0.99) | 0.99 (0.98, 1.00) | 0.95 (0.91, 0.98) | 0.91 (0.85, 0.95) | 0.87 (0.81, 0.92) |
| Category 2 (WF) | 0.98 (0.95, 1.00) | 0.97 (0.93, 1.00) | 0.98 (0.95, 1.00) | 0.94 (0.89, 0.98) | 0.97 (0.93, 1.00) | 0.93 (0.88, 0.97) | 0.74 (0.66, 0.82) | 0.79 (0.71, 0.87) |
| Category 3 (HRS) | 0.92 (0.80, 1.00) | 0.83 (0.68, 0.96) | 0.83 (0.68, 0.96) | 0.83 (0.68, 0.96) | 0.83 (0.68, 0.96) | 0.89 (0.76, 1.00) | 0.66 (0.50, 0.81) | 0.62 (0.46, 0.78) |

b.

| Cyst Risk Category | Risk Categorization Recall with 95% Confidence Interval | | | | | | | |
|---|---|---|---|---|---|---|---|---|
| | GPT-CoT | DeepSeek-FT-CoT-SC | DeepSeek-FT-CoT | LLAMA-FT-CoT | LLAMA-FT | GPT | DeepSeek | LLAMA |
| Category 1 (Low Risk) | 0.99 (0.97, 1.00) | 0.96 (0.93, 0.99) | 0.97 (0.94, 0.99) | 0.94 (0.90, 0.98) | 0.96 (0.93, 0.99) | 0.95 (0.91, 0.98) | 0.73 (0.66, 0.80) | 0.80 (0.74, 0.86) |
| Category 2 (WF) | 0.97 (0.93, 1.00) | 0.97 (0.93, 1.00) | 0.96 (0.92, 0.99) | 0.95 (0.91, 0.99) | 0.98 (0.95, 1.00) | 0.90 (0.84, 0.96) | 0.86 (0.79, 0.93) | 0.79 (0.71, 0.87) |
| Category 3 (HRS) | 0.96 (0.86, 1.00) | 0.96 (0.87, 1.00) | 0.96 (0.87, 1.00) | 0.96 (0.87, 1.00) | 0.96 (0.87, 1.00) | 1.00 (1.00, 1.00) | 1.00 (1.00, 1.00) | 0.92 (0.79, 1.00) |